# A FAST DECISION TECHNIQUE FOR HIERARCHICAL HOUGH TRANSFORM FOR LINE DETECTION

Chandan Singh[1] and Nitin Bhatia[2]

[1]Professor and Head, Department of Computer Science, Punjabi University, Patiala-147002, India. chandan@pbi.ac.in
[2]Lecturer, Department of Computer Science, DAV College, Jalandhar-144008, India. n_bhatia78@yahoo.com

**ABSTRACT**

*Many techniques have been proposed to speedup the performance of classic Hough Transform. These techniques are primarily based on converting the voting procedure to a hierarchy based voting method. These methods use approximate decision-making process. In this paper, we propose a fast decision making process that enhances the speed and reduces the space requirements. Experimental results demonstrate that the proposed algorithm is much faster than a similar Fast Hough Transform.*

**KEY WORDS**

Edge Detection, Line Detection, Hough Transform, Hierarchical Hough Transform.

## 1.Introduction

Edge detection is one of the most frequent and important tasks of image processing. Line detection plays major role in detection of edges. Hough Transform [1] is accepted as an important tool to perform this step. It has also been extended to detect other shapes such as circles, ellipses, etc [2-3]. There is enormous amount of literature available about this algorithm.

Hough Transform is a special case of Radon Transform [7]. Radon Transform and Inverse Radon Transform have been widely used to detect lines in digital images [4-7]. Recently a new method has been proposed to detect strips of lines instead of thin lines using Radon transform [5]. Inverse Radon transform has been used to detect lines but only in the presence of prior information about lines using dictionary of lines [7].

The problem with Hough Transform algorithm is that it uses ρ (the perpendicular distance of line from origin) and θ (the angle made by normal to this line with the positive direction about x-axis). This makes it time consuming. If m (slope) and b (y-intercept) are used instead of ρ and θ, then the parameter space becomes too big to handle. Solution to this problem is also available [8]. The lines are divided into two different sets, one with absolute value of slope less than or equal to 1 and another with absolute value of slope greater than 1. The lines with slope greater than 1 are handled by inverting the roles of x and y-axes, consequently making the absolute value of slope less than or equal to 1. All these algorithms require an edge detector to be applied before starting the process.

The proposed algorithm is based on the slope-intercept technique. Each edge point maps to a line in the parameter space of m and b. In hierarchical Hough transform algorithm [9], the parameter space is divided into four sub quadrants if number of lines passing through that region exceeds a threshold. To decide the number of lines passing through a region, each line votes for that region. If the number of votes secured by a region crosses a pre defined threshold value then that region is subdivided otherwise that region is stopped from further subdivision. The problem of voting is simplified to minimize the computations and reduce the frequency of a line participating in polling.

In the Fast Hough Transform (FHT) [9], the detection of lines is carried out through a hierarchical approximation to the solution, which enhances the speed. The problems with this algorithm are errors due to floating point operations and false voting by some of the lines.

This paper presents a new algorithm for making the decision as to whether a line must vote for a region or not. The algorithm is faster than the algorithm presented in [9] because it uses bit-wise operations, whereas the algorithm in [9] uses approximations involving floating point operations in deciding the membership of a line to a region. Accordingly, as given in Fig. 1, line 1 passes through the region whereas 2 and 3 do not. The algorithm given in [9] assumes that line 2 also passes through the region. Hence, line 2 is a candidate for voting even for the subdivided regions.

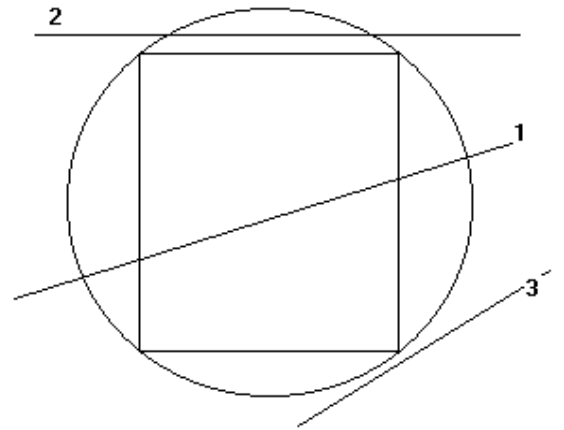

**Fig. 1 Different lines treated by algorithm in [9].**

The algorithm presented in this paper does not consider this approximation as it uses an exact method to decide the membership of a line to a region. Hence, lines such as line 2 of Fig. 1 do not vote for the region and are not the candidates for further subdivisions. This reduces the amount of computation resulting in faster results.

## 2. The Method

In section I, we pointed out the problem with FHT given in [9], which uses approximations involving floating point operations and unwanted voting. It calculates the perpendicular distance to the line from the center of the region and compares it with the radius of the circle circumscribing the region. This is an approximation. The lines falling in category 2, as explained in section I, are being considered to be in the region but actually these are not. The proposed algorithm decides the same thing depending upon whether a line actually passes through the region or not. We assign a 4-bit code to each of the lines with respect to each of the region. The 4-bit code contains the information whether a line belongs to a region or not. The lines falling in category 2 are rejected automatically reducing the calculations because these lines will never be entertained. Moreover, the 4-bit code is calculated and handled with bit-wise operations making the operations much faster as compared to FHT [9]. Figure 2 describes how various lines get different bit codes depending upon the edges these lines cross. We consider the four corner points in the order of North East, North West, South West and South East. The left most bit pertains to the point North East and the right most bit to the point South East. The process of assigning the bit values is given as follows:

a). We initialize all bits to zero.
b). A bit is 1, if the corner point is below the line.

Accordingly the bit values for line 1 in Fig. 2 are 1, 1, 0 and 1, because the corner North East is below the line, North West is above the line and South East and South West are below the line. We say that a point $(x_0, y_0)$ is above the line, if the value of the expression

$$F(x_0, y_0) = y_0 - m \cdot x_0 - b$$

is positive or zero. Otherwise, the point will be below the line. All the lines here have absolute slope less than or equal to 1.

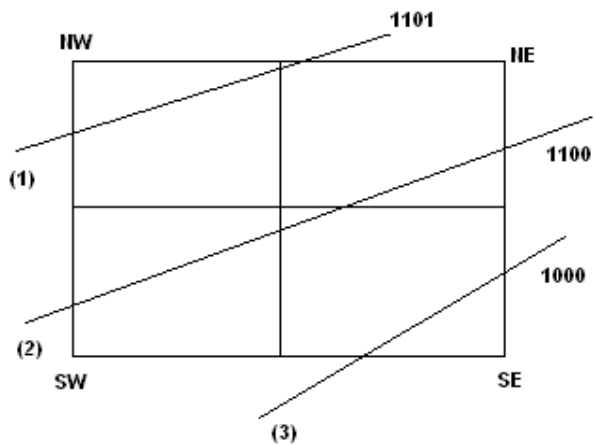

**Fig. 2 Different lines treated by proposed algorithm.**

It may also be noted that, the presence vectors and the distance vectors taken by FHT consume lot of memory in [9]. These two vectors are not required in the proposed algorithm as the 4-bit code fulfils all requirements.

## 3. The Algorithm

The Hough Transform maps edge points in the image to lines in Hough parameter space. The points in the Hough parameter space getting sufficient number of votes, appearing as peaks, are detected. These points in Hough space, in turn, provide true lines in image space.

The algorithm processes two sets of edge points. One called *ALPHA* with edge points belonging to lines with absolute slope less than or equal to 1 and other called *BETA* with rest of the edge points. These edge points can be put to these sets while the Sobel's gradient operator is being applied to the image. The algorithm first processes the set of edge points belonging to ALPHA and then repeats the same process on the set of edge points belonging to BETA but with reversing the roles of co-ordinates for each edge point. For a point $(x_0, y_0)$ in BETA, we assume the point to be $(y_0, x_0)$. By doing so, the absolute value of the slope becomes less or equal to 1. The algorithm starts with assuming the region to be the complete Hough parameter space. Obviously, each line in Hough parameter space corresponding to each edge point in the set passes through this region. We call it level 0. The algorithm divides the region into four sub quadrants and each such division takes the process to next level. The algorithm is, therefore, implemented using a quad tree data structure each node of which represents a region. The purpose of the algorithm is to decide about the votes, which each region gets in the Hough space. If a region does not get enough votes required to produce a true line in the image then further subdivision of the region is stopped. This enhances the speed of the algorithm.

To decide if a line passes through a region, FHT in [9] proposes a method that uses an approximation. It compares the shortest distance of line from the centre of the region and the radius of circum circle of the region to decide if the line passes through within the region or not. In this process, lines passing outside the region but passing inside the circum circle also vote for that region. This false voting scheme reduces the accuracy of line detection process. Moreover, the decision process uses floating-point calculation including square root operations, which makes the algorithm slow. To decide if a line passes through a region or not, the proposed algorithm uses a faster technique. It assigns a 4-bit code to each line at each level for each quadrant. This 4-bit code is assigned values according to the location of the line with respect to all four corners of the region. If a corner point lies above the line the corresponding bit is assigned the value 0 otherwise it is assigned value 1 as explained in section II. While dividing the regions into subsequent four quadrants the size of the region may reduce to a single pixel provided that the region gets sufficient votes required to produce a true line in the image. This happens at level $\lceil \log_2 N \rceil$ for an image of size $N \times N$. At this level, the region that is of the size one pixel provides the coordinates of Hough parameter space. These parameters represent the peaks in Hough space and result in true line in image space.

The pseudo code for the algorithm is given as:
Begin
1  level ← 0
    level.vote ← 0
2  Repeat steps 3 through 5 for all edgepoints

```
3    code ← getcode()    // explained later
4    level.linecode ← code
5    if code ≠ 0000 and code ≠ 1111 then
        // vote for this region
            level.vote ← level.vote+1
     endif
6    level ← level+1
7    level.quad ← 0
8    Repeat steps 9 through 12 until level is 0
9        level.vote ← 0
10       level.quad ← level.quad+1
         // go to next sub quadrant of region
11       call the function calculate_center_for_subregion()
         //explained later
12       if level.quad <= 4 then
         //only four sub quadrants can be there
         a) Repeat step 14 for all edgepoints
         b) if (level-1).linecode ≠ 0000 and
            (level-1).linecode ≠ 1111 then
         // if the line passes through the parent region
         // then decide about the child regions
             i. if level.quad = 1 and (level-1).linecode = 0100 then
                 // exclude the first sub quadrant if code for
                 // parent region is 0100
                     level.linecode ← 0000
                     goto step 13
                 endif
             ii. Repeat step i. for other quadrants with
             appropriate values of linecode.
             iii. code ← getcode()
             iv. level.linecode ← code
             v. if code ≠ 0000 and code ≠ 1111 then
                 // point votes for this region
                     level.vote ← level.vote+1
                 endif
             else
                 level.linecode ← 0000
         endif
13   if level.vote > THRESHOLD then
// if votes cross the threshold value then check for the level.
// if the region reduces to one pixel then the points need
// be plotted else increase value of level, i.e., divide
// the region further into sub quadrants.
         if level = ⌈log₂N⌉ then
             call solution()    // explained later
         else
             level ← level +1
             level.quad ← 0
         endif
       endif
     else
       level ← level-1
     endif
End
```

In the above algorithm, *level* 0 is the region covered by the original image. As we subdivide the region into four quadrants, we get four sub quadrants of the *level* 0. These sub quadrants are referred to as *quad1, quad2, quad3* and *quad4* which represent *NorthEast, NorthWest, SouthWest* and *SouthEast* quadrants, respectively. Each *quad* here falls under *level* 1. As *level* 1 is subdivided, we get four more quadrants for each *quad* and so on. Each edge point maps to a line in the working parameter space. Therefore, each line may or may not pass through the region under consideration. Each line gets a 4-bit *code*, which shows whether the line passes through the region or not. The variable *level.linecode* gives the code for line with respect to the *level*. The variable *level.vote* is a measure of the number of lines passing through the region of that *level*. When a *level* secures votes more than a predefined THRESHOLD value, we assume that the *level* requires subdivision. Any *level* not meeting the criteria of THRESHOLD value is subdivided and the value of *level* is increased by one. For an image of size N × N, *level* ⌈log₂N⌉ is of size one pixel. Therefore, we stop the process.

The function *getcode()* is responsible for providing *code* to each of the edgepoints by judging the location of the line with respect to the corners of the sub region. The *getcode()* function is given as:

Code ← 0000
If *NorthEast* corner of the region is below the line then
    Bit 1 is set to 1
If *NorthWest* corner of the region is below the line then
    Bit 2 is set to 1
If *SouthWest* corner of the region is below the line then
    Bit 3 is set to 1
If *SouthEast* corner of the region is below the line then
    Bit 4 is set to 1

The procedure *calculate_center_for_subregion()* is responsible for calculating the centers of the subsequent sub regions being inspected by the algorithm. Centre co-ordinates of child region can easily be calculated if we have the size of the parent region in advance. For example, the coordinates of center of *North East* subregion will be:

    level.x ← (level-1).x + N/2$^{level}$
    level.y ← (level-1).y + N/2$^{level}$

Similarly, for *South West* sub region, center co-ordinates will be

    level.x ← (level-1).x - N/2$^{level}$
    level.y ← (level-1).y - N/2$^{level}$

The procedure *solution()* is given by:
Begin
1. level ← ⌈log₂N⌉
2. Repeat step 3 for all edgepoints
3. If level.linecode ≠ 0000 and level.linecode ≠ 1111 then
       a) plot the edgepoint
       b) for k = 0 through level
          // if an edge point has been plotted then
          // exclude the edge point from further calculations
            k.linecode ← 0000
       end for
   endif
End

The above subroutine provides the solution points as the true edge points and also removes them from taking part in subsequent computations. All the points selected for solution are reflected in the output image. The above procedure has been implemented in VC++ 6.0 under Windows. Interested readers may request the authors for the source code.

## 4. Evaluation

The proposed algorithm is evaluated on the basis of the following aspects: accuracy, memory requirements and time complexity.

### 4.1 Accuracy

Apart from ruling out the approximation done by FHT in [9] and reducing the memory and time requirements, the proposed algorithm produces better results than those provided by FHT in [9]. We have chosen 256× 256 pixel gray scale image in TIFF format. FHT in [9] uses vectors to store coefficients, which require floating point calculations. Due to these floating point calculations, the FHT in [9] either creates some new pixels or misses some of them near the edges in the image. The results of various algorithms are shown here on different images. We took various values of THRESHOLD and observed that a value around 40 gives acceptable results. Figures 3 through 5 provide the visual results of the proposed algorithm and the results of FHT [9]. The original images have also been depicted in the figures. It can be observed visually that the proposed algorithm provides better results.

### 4.2 Memory Requirements

Assume that the number of edge points in the image is E. The proposed algorithm requires a memory space of the size:

$$M = \tfrac{1}{2}E(\lceil \log_2 N \rceil + 1) \text{ words}$$

because there will be a maximum of $\lceil \log_2 N \rceil + 1$ levels. The FHT in [9] requires $E(\lceil \log_2 N \rceil + 1)$ words+$E(\lceil \log_2 N \rceil + 1)$ bits.

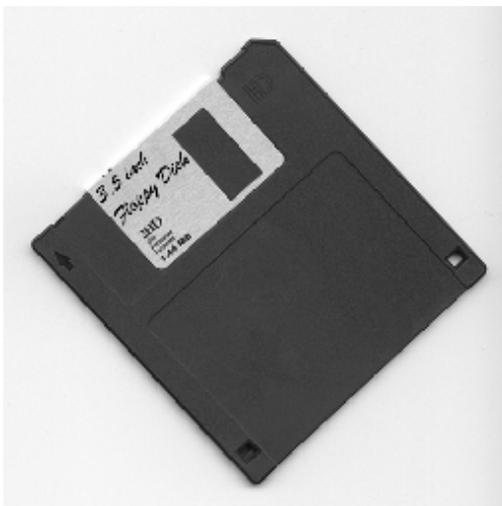

(a)

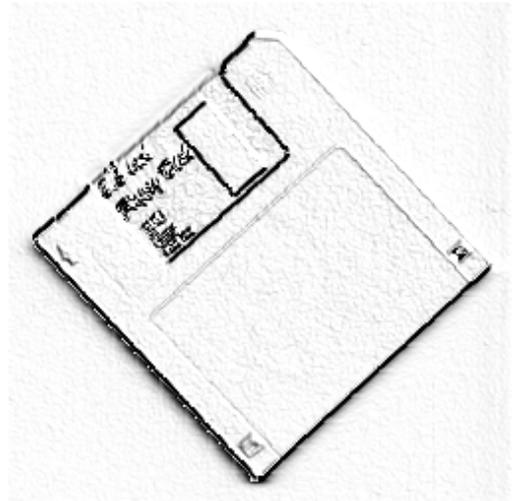

(b)

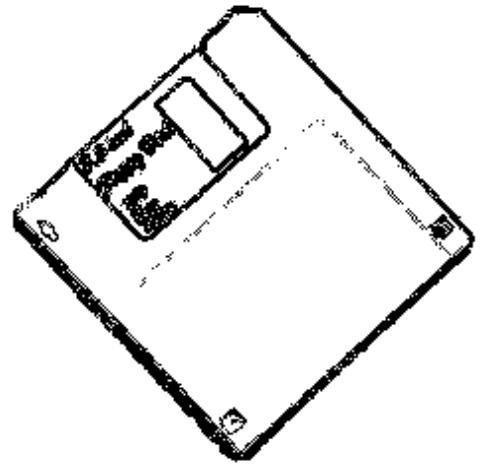

(c)

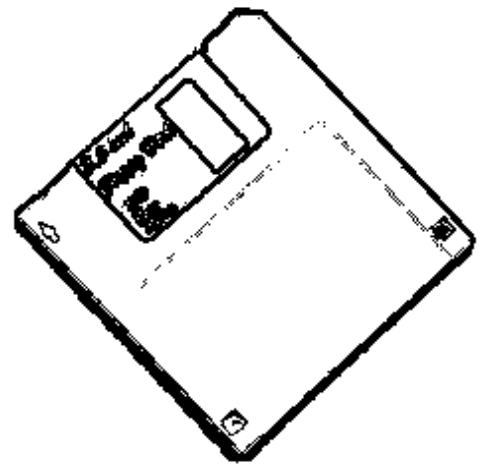

(d)

**Fig. 3. (a) Sample image 1; (b) Result after applying Sobel's operator; (c) Line detection using FHT in [9]; (d) Result of applying proposed algorithm.**

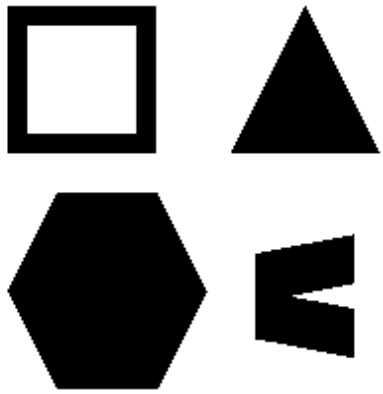

(a)

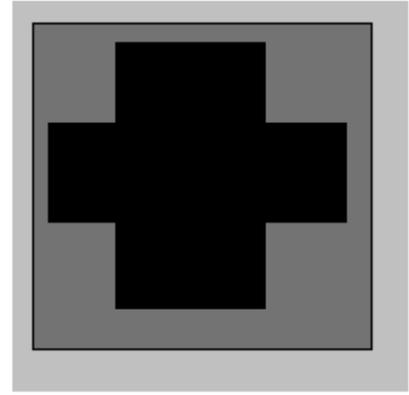

(a)

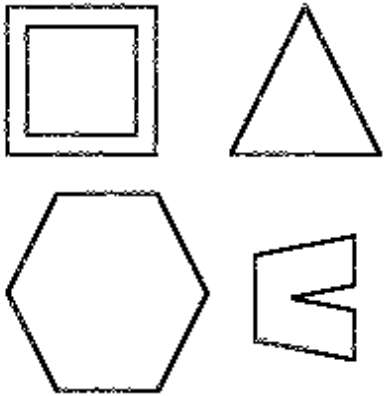

(b)

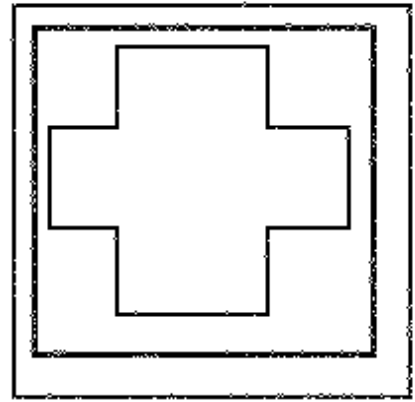

(b)

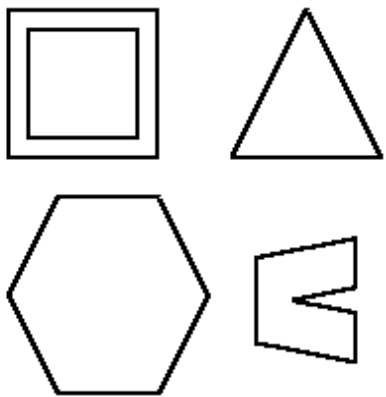

(c)

Fig. 4. (a) Sample image 2; (b) Result of FHT in [9]; (c) Result of proposed algorithm.

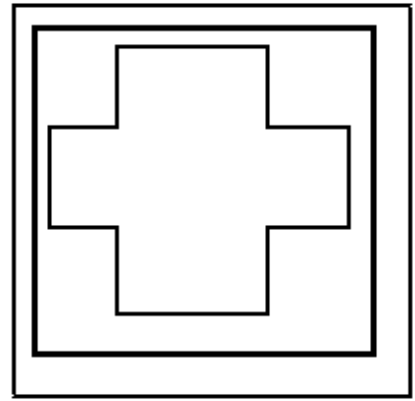

(c)

Fig. 5. (a) Sample image 3; (b) Result of FHT in [9]; (c) Result of proposed algorithm.

## 4.3 Time Complexity

The time required by the traditional Hough Transform is of the order of E*N where the number of edge points in the image is E. The FHT in [9] requires 3*E floating point operations to calculate the coefficients. In addition it requires one square root operation for each edge point. Apart from this, it needs extra time to calculate the distance. The proposed algorithm requires much less computation time as the average case of this algorithm is better than the best case of FHT in [9]. The reason is that we do not undergo any floating point operation as are done in case of FHT in [9] to compute the required coefficients. Also the approximation, as explained in section I, increase a lot of computations if majority of the lines fall into category 2. This case is ruled out in the proposed algorithm. Even in the worst case, the number of times the proposed algorithm computes the 4-bit code is

$$4E(\lceil \log_2 N \rceil + 1)$$

which of course requires bit wise operations. The proposed algorithm does not require any calculation of coefficients or distance as is done by FHT [9].

| Size | Image | | Fig. 3(a) | Fig. 4(a) | Fig. 5(a) |
|---|---|---|---|---|---|
| 32X32 | Number of votes | FHT of [9] | 3391 | 4577 | 6081 |
| | | Proposed algorithm | 1817 | 2387 | 3277 |
| | Execution time (in sec) | FHT of [9] | 4.625 | 6.125 | 8.375 |
| | | Proposed algorithm | 0.031 | 0.031 | 0.046 |
| 64X64 | Number of votes | FHT of [9] | 10424 | 13537 | 18281 |
| | | Proposed algorithm | 4551 | 5945 | 8051 |
| | Execution time (in sec) | FHT of [9] | 15.797 | 18.015 | 23.827 |
| | | Proposed algorithm | 0.062 | 0.078 | 0.109 |
| 128X128 | Number of votes | FHT of [9] | 71315 | 70946 | 119756 |
| | | Proposed algorithm | 15759 | 15691 | 26513 |
| | Execution time (in sec) | FHT of [9] | 110.387 | 111.764 | 181.243 |
| | | Proposed algorithm | 0.157 | 0.179 | 0.235 |
| 256X256 | Number of votes | FHT of [9] | 393553 | 219320 | 342795 |
| | | Proposed algorithm | 51420 | 29112 | 45457 |
| | Execution time (in sec) | FHT of [9] | 648.689 | 359.421 | 573.092 |
| | | Proposed algorithm | 0.484 | 0.25 | 0.306 |

Table 1. Comparison of number of votes (algorithms executed once) and execution time (in seconds) (algorithms executed 1000 times) between FHT of [9] and the proposed algorithm.

We have conducted a number of experiments to find the execution time required by both the algorithms. The programs were executed on a Pentium-IV PC having 3 GHz CPU and 256 MB memory. All the three images have been analyzed for edge detection with four different sizes: $32 \times 32$, $64 \times 64$, $128 \times 128$ and $256 \times 256$. Since the absolute time taken by the proposed algorithm is very less (it shows 0 millisecond as execution time for an image of size $32 \times 32$), in order to highlight the difference, we have executed both the algorithms under loops running for 1000 times. The number votes voted by and time taken (in seconds) by both the algorithms have been given in Table 1. The ratio of execution time, $C_t$, and the ratio of total number of votes, $C_v$, have been taken as

$$C_t = \frac{\text{Execution time required by FHT [9]}}{\text{Execution time required by proposed algorithm}}$$

$$C_v = \frac{\text{Number of votes voted by FHT [9]}}{\text{Number of votes voted by proposed algorithm}}$$

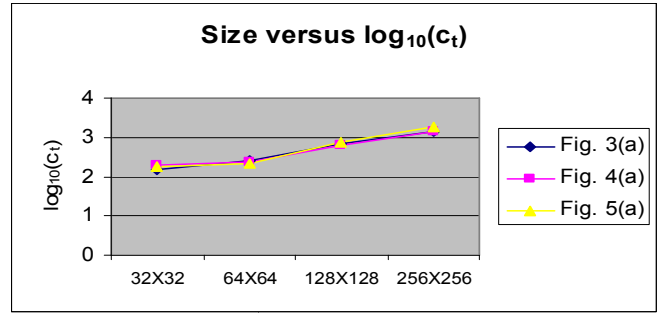

Fig. 6. Size versus $\log_{10}(C_t)$ (algorithms executed 1000 times)

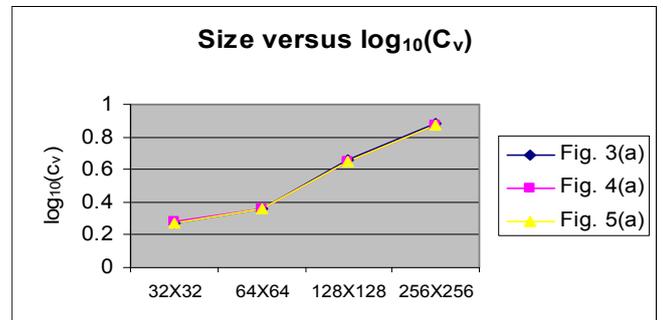

Fig. 7. Size versus $\log_{10}(C_v)$ (algorithms executed 1000 times)

It is clear that these ratios have exponential nature with respect to the size of the image. The logarithms of these factors have been plotted in Fig. 6 and Fig. 7 which clearly demonstrate that the proposed algorithm is much faster. This improvement in execution is expected as the proposed algorithm uses integer arithmetic and the number of votes voted by our algorithm is less by a factor between 7 and 8 for $256 \times 256$ images. The major factor, however, is the floating point and square root operations involved in [9]. We believe that the implementation of the algorithm may affect the results but the deviations would not be significant.

## Conclusion

Hough transform is a widely researched algorithm. The hierarchical implementation of this algorithm provides faster and accurate results. We have presented a new algorithm for line detection in images, which is a modified version of some of the already available algorithms. The proposed algorithm is efficient in terms of speed and memory requirements and produces better results. Experimental results show that the proposed algorithm is much faster. Like other Hough Transform based algorithms, it has the features of parallelization and detection capability of other shapes such as circles, ellipses, rectangles etc. for which the authors are carrying out research.